\documentclass[11pt,a4paper]{article}
\usepackage[hyperref]{acl2017}
\usepackage{times}
\usepackage{latexsym}
\usepackage{graphicx}
\usepackage{amsmath}

\usepackage{url}

\aclfinalcopy 
\def\aclpaperid{22} 

\newcommand{\jcs}[1]{\textcolor{green}{\bf \small [JCS-- #1]}}

\title{Domain Aware Neural Dialog System}

\author{Sajal Choudhary 
 \qquad
  Prerna Srivastava 
   \qquad Lyle Ungar 
  \qquad Jo\~ao Sedoc \\
         Computer \& Information Science\\
{University of Pennsylvania} \\
{\tt sajal, prernasr, ungar, joao @cis.upenn.edu} \\
 }

\begin{document}
\maketitle
\begin{abstract}
  
We investigate the task of building a domain aware chat system which generates intelligent responses in a conversation comprising of different domains. The domain in this case is the topic or theme of the conversation. To achieve this, we present \textit{DOM-Seq2Seq}, a domain aware neural network model based on the novel technique of using domain-targeted sequence-to-sequence models~\cite{DBLP:journals/corr/SutskeverVL14} and a domain classifier. The model captures features from current utterance and domains of the previous utterances to facilitate the formation of relevant responses. We evaluate our model on automatic metrics and compare our performance with the Seq2Seq model. 
\end{abstract}

\section{Introduction} 
With the advent of personal assistants such as Siri and Alexa, there has been a renewed focus on dialog systems, specifically those which can hold open-domain conversations. Readily available conversations from social media such as Reddit \footnote{https://www.reddit.com/} have facilitated the use of data driven models to generate open domain responses. A recurrent neural network-based sequence-to-sequence model has been successfully applied to generating responses for conversational tasks ~\cite{DBLP:journals/corr/VinyalsL15}. 

Though these models generate responses that are fluent and natural, they often fail to produce domain-specific responses since they are trained on a single general data-set. Moreover, these models tend to capture information solely based on previous words in a single utterance and fail to acknowledge switching of domains in a multi-turn conversation. Recent research has aimed at solving these problems and has shown promising results. Topic-aware neural networks ~\cite{DBLP:conf/aaai/XingWWLHZM17} seek to incorporate topic information in their model to generate responses relevant to the topic words in the query. However, the context inherent in the previous utterances is not taken into account. On the other hand, hierarchical neural networks~\cite{DBLP:conf/aaai/SerbanSBCP16,DBLP:conf/aaai/SerbanSLCPCB17} model the interactive structure of a conversation in order to output more contextual responses. Though these models try to capture the context of previous utterances in a conversation, the responses generated might lack domain-specific relevance.  

\begin{figure}
    \centering
    \includegraphics[width=3in]{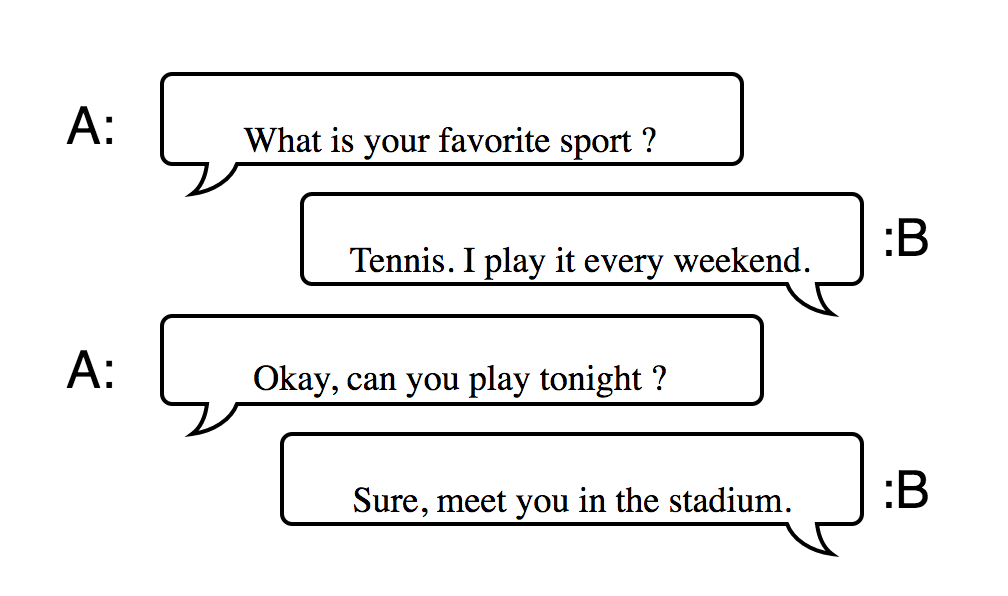}
    \caption{Two users(A and B) conversing with each other. At second turn, B still remembers the domain of the conversation and hence answers appropriately in the same domain.}
    \label{fig:fig_1}
\end{figure}

 We hypothesize that each utterance in a conversation belongs to a domain and that during a conversation, participants tend to switch between domains. Moreover, sometimes the domain of an utterance might be ambiguous, but incorporating the previous domains of the conversation might be  helpful in inferring the precise domain. For instance in Figure \ref{fig:fig_1}, the first utterance by speaker A lies in the \textit{sports} domain. The domain for A's second utterance is ambiguous and could lie in either \textit{sports} or \textit{music}. Since the previous domains referenced by the participants is \textit{sports}, it is highly probable that the exchange is about playing ``tennis" rather than playing a song named ``tonight". When B is a chatbot, this utterance might be misinterpreted as a request for playing a song. We address the above problems by modelling a conversation and taking into account the domain of the current utterance as well as domains of the previous utterances.

In this paper, we describe a novel domain aware dialog system, DOM-Seq2Seq, consisting of three main components: 1) Domain Classifier, 2) Domain specific response generators (Seq2Seq),  and 3) Re-ranker to combine responses of the above two models. 

\section{Model}
\subsection{Overview}
\begin{figure}[h]
    \centering
    \includegraphics[width=3in]{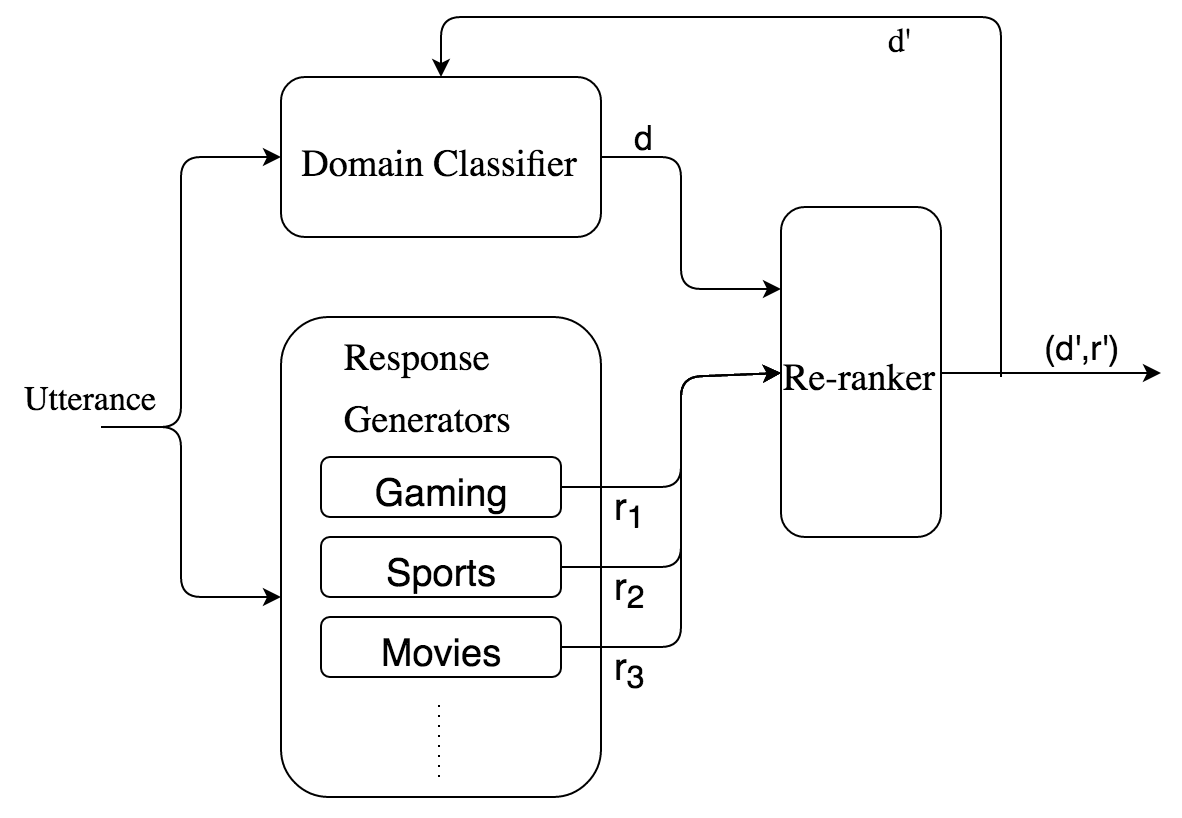}
    \caption{High level overview of the DOM-Seq2Seq conversational model. The utterance is fed into two separate models: Domain Classifier and Response generators. Their output is fed to re-ranker. Re-ranker outputs the final predicted domain($d'$) and response($r'$). $d'$ acts as a feedback to the domain-classifier. }
    \label{fig:fig_2}
\end{figure}

In order to maintain a smooth conversation within the domain as well as during switching of domains in a conversation, we built a model consisting of a domain classifier and a set of response generators. The overview of the model is depicted in Figure \ref{fig:fig_2}. An utterance is fed into the domain classifier as well as multiple response generators which are trained separately on domain-specific data. The output domain ($d$) predicted by the domain classifier and all the generated responses ($r_1$, $r_2$, $r_3$) are then fed to the re-ranker. Re-ranker outputs the final predicted domain ($d'$) and the appropriate response ($r'$). The final predicted domain ($d'$) is fed back to the domain classifier to be used in subsequent predictions.

\subsection{Domain classifier} \label{domain classifier}
Domain classification predicts the domain of an utterance. Firstly, we used a tf-idf based supervised SVM to capture utterance level word features. Subsequently, we employed two ways to combine predicted domain from SVM and previous set of domains in the conversation.

\subsubsection{Ensemble based Domain Classifier} \label{ensemble domain class}
We applied a logistic regression model to capture the sequence of domains in the conversation. While training the model, three previous actual domains ($d_{t-1}$, $d_{t-2}$, $d_{t-3}$) are taken into consideration, where $d_{t-1}$ is the domain of the previous utterance and so on. Along with this, the domain predicted by the SVM model ($d_{t}^{SVM}$) is also included as a feature. Hence, this model optimizes the cost function,
\begin{equation}    
p(y|d_{t-1}, d_{t-2}, d_{t-3}, d_{t}^{SVM})
\end{equation}
where y is the actual label (domain) of the utterance in the conversation.

\subsubsection{RNN based Domain Classifier}
\begin{figure}[h!]
    \includegraphics[width=3in]{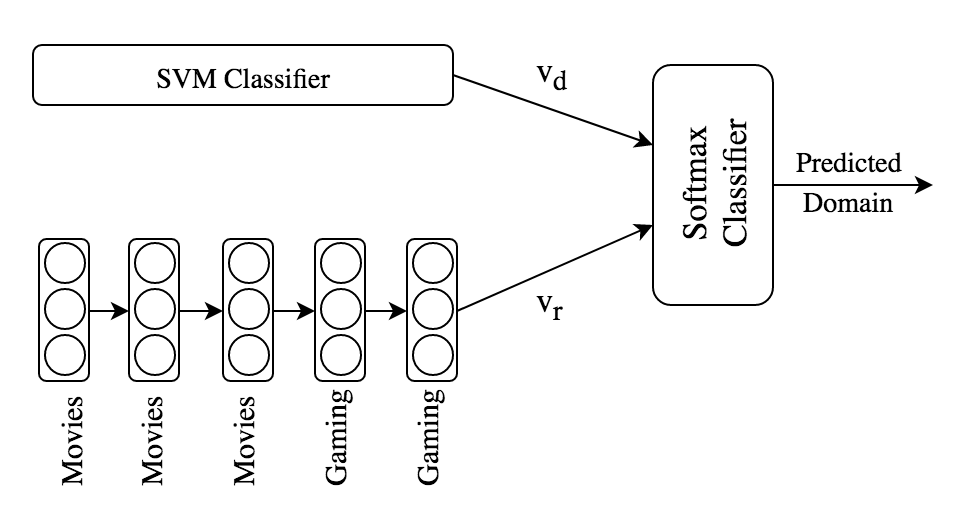}
    \caption{RNN Based Domain Classifier: The vector representation of the output from SVM tf-idf ($v_d$) is concatenated with last the hidden state of RNN. These are then fed to softmax classifier. }
    \label{fig:fig_3}
\end{figure}

Our previous approach (\ref{ensemble domain class}) is based on the assumption that the last three domains in the conversation are sufficient to predict the current domain. However, capturing long term dependencies on domains is more beneficial for this task instead of relying only on the last few domains. The best way to capture this information is to use an RNN model with one hot input vector representation of subsequent domains. This is depicted in Figure \ref{fig:fig_3}.
\noindent The state at time step t for an RNN is represented as  $s_t = f(d_t,s_{t-1})$ where $s_t$ is the hidden state of the RNN at $t^{th}$ time step and $d_t$ is the domain at the same time step. Thus, at each time step, the state modifies itself based on its previous state and the current input. In our case, RNN is fed with the sequence of domains. The last hidden state ($v_r$) captures all the necessary information about the sequence of domains. The SVM model's prediction is also represented in a vector form ($v_d$). These two vectors are concatenated and fed to a softmax classifier to get a probability distribution over the domains as in Equation \ref{prob}.
\begin{equation} \label{prob}
p(d_t|d_{<t},v_d) = Softmax(W*[v_r||v_d]+b)
\end{equation}
\subsection{Response Generators}
For generating the output response, we used an LSTM based Seq2Seq model with attention mechanism~\cite{DBLP:journals/corr/BahdanauCB14}. Seq2Seq consists of two recurrent neural networks: encoder and decoder. An encoder compresses the input sentence into a fixed length vector, which is then passed to the decoder. The decoder then generates the output response one word at a time. 
Thus a candidate response is output for each domain. During testing phase, the output of the decoder at time step t is fed to the input of decoder at time step t+1; it outputs the logit at each time step. We extract the output logit $l$ from the last time step of the decoder and apply sigmoid function to limit its value between 0 and 1. The value of $ \sigma(l) = \frac{1}{1+e^{-l}}$ is used in the re-ranker stage to combine it with the output of the domain classifier. 

\subsection{Re-ranker}
The output scores from the the generators and the domain classifier are fed into a re-ranker which implements soft classification to figure out the best response from the set of candidate responses. Equation \ref{eq5} returns the index of the response with the highest multiplied score.  So,
\begin{equation} \label{eq5}
\textrm{final response index} =  \operatorname{arg\,max}_{i} \{ p(d_i) * p(r_i)\},
\end{equation}
where $p(d_i)$ is the probability that the utterance belongs to $i^{th}$  domain and $p(r_i)$ is the $i^{th}$ domain's decoder probability which was calculated from its output logit.

\section{Experiments}
In this section, we describe the experiments performed for our proposed models. We present the quantitative results of our evaluations and compare them with the those of the Seq2Seq model. We also put forward some qualitative examples of conversation snippets tested on our models. Our experiments were restricted to three domains namely \textit{Movies}, \textit{Gaming} and \textit{Out of Domain} to aid in data collection and evaluation. We introduced \textit{Out Of Domain} to manage utterances that do not lie in either \textit{Movies} or \textit{Gaming}. 

\subsection{Data Sets}
We collected data from the social discussion website Reddit for training our models. We gathered 1,334,119, and 542,594  query-response pairs from the \textit{Movies} and \textit{Gaming} Reddit categories respectively for training two Seq2Seq models. Furthermore, for responding to \textit{Out of Domain} queries, we trained Seq2Seq using 301,812 query-response pairs from twitter. 

The SVM classifier for the domain classifier was trained and tested on 544,630 posts from Reddit categories of \textit{Movies}, \textit{Gaming} and tweets from Twitter combined. We made sure that there was no overlap with the data used for training of Seq2Seq models.

\begin{table}[t]
\centering
\begin{tabular}{  p{7cm}  }
  \textbf{Conversation}   \\
\hline
\vspace{0.05cm}
\textbf{User}: hi how are you \newline \textbf{Seq2seq}: morning bro \newline \textbf{DOM-Seq2Seq}: hello how are you  \\ \\
\textbf{User}: what games interest you \newline \textbf{Seq2seq}: Super Life \newline \textbf{DOM-Seq2Seq}: Sonic 0, The Walking \\ \\
\textbf{User}: want to play tonight \newline \textbf{Seq2seq}: lol this is the best \newline \textbf{DOM-Seq2Seq}: I wonder what you're doing \\  \\
\textbf{User}: do you like watching movies \newline \textbf{Seq2seq}: [no response] \newline \textbf{DOM-Seq2Seq}: I'm a fan of them \\ \\
\textbf{User}: the x men movies are good \newline \textbf{Seq2seq}: i love this \newline \textbf{DOM-Seq2Seq}: i thought i saw them \\ \\
\textbf{User}: do you want to watch a movie \newline \textbf{Seq2seq}: i' m not sure he's a great \newline \textbf{DOM-Seq2Seq}: I have a feeling I'd watch them in the theater \\
\hline
\end{tabular}
\caption{Qualitative comparisons of responses from the Seq2Seq baseline and DOM-Seq2Seq.}
\label{tab:examples}  
\end{table}

\subsubsection*{Tagging Conversational Data with Domains}
Conversations extracted from Reddit IAma were used as data for the model. Reddit IAma threads specially suited our purposes, as they are already tagged with a flair, which is essentially the category of the conversation. We found that participants often switch categories during the conversations as these are of the type ``ask me anything". Each utterance of conversation was labelled automatically. Latent Dirichlet Allocation~\cite{Blei:2003:LDA:944919.944937} was used to learn topics from the conversation and to label each utterance with a topic corresponding to a domain if the topic proportion for that domain was above the threshold of $0.5$. We replaced any domains outside of \textit{Movies} and \textit{Gaming} by \textit{Out Of Domain}. Finally, we weighted the previous domains with exponentially decaying weights and took an average of these to incorporate the effect of previous domains. We amassed 947 conversations.
\subsection{Experimental Setup} 
\subsubsection*{Response Generators}
We used a Seq2Seq model with 3 layers of LSTM each consisting of 1024 hidden units as encoder and decoder. The vocabulary size for the Seq2Seq for \textit{Movies}, \textit{Gaming} and \textit{Out Of Domain} was kept at 253,468, 85,312 and 40,000 words, respectively. These values were decided after filtering out the words which occurred once in their respective domains. Early stopping was used based on perplexity values on validation set.
\subsubsection*{RNN based Domain Classifier}
A single layer RNN architecture with 8 hidden units was employed for the domain classifier and trained for 50 epochs.  

\subsection{Evaluation}
We evaluated our models along with the Seq2Seq model on two metrics 1) Domain Classification Accuracy of the domain classifier and 2) Word Embedding Greedy Match~\cite{DBLP:conf/emnlp/LiuLSNCP16}. The GloVe word embedding metrics~\cite{pennington2014glove} were used to find the similarity between generated response and the ground truth. This metric signifies how semantically similar the generated response is to the true response, and by extension, related to the query. The results for each of the models are specified in the Table \ref{tab:eval}.

\section{Analysis}
DOM-Seq2seq models outperform Seq2Seq on the Word Embedding Greedy Match metric showing that response from these models are more suitable and similar to the ground truth. En-DOM-Seq2Seq surpasses RNN-DOM-Seq2Seq with respect to domain classification accuracy, as conversations in our data set are relatively short.  Table \ref{tab:examples} shows the responses generated by Seq2Seq and DOM-Seq2Seq. Both of the DOM-Seq2Seq models  predicted the same domain in our example conversation. Thus, for brevity, we showed only one response to compare it against Seq2Seq. The responses by DOM-Seq2Seq models are more informative. The response to query ``do you want to watch a movie" from DOM-Seq2Seq is more pertinent than the one from Seq2Seq.

\begin{table}[h]
\centering
\begin{tabular}{ p{2cm} | p{2cm} | p{2cm}  }
{\bf \ \ \ \ \ \ \ \ \ \ \ \ \ \ \ \ \ \ \ \ \ \ \ \ \ \ \ \ \ \ \ \ \ \ \ \ \ \ \  \ \ \ \ \ Model} 
& \textbf{Domain Classifier Accuracy}  & \textbf{Word \ \ \  \ \ \ Embedding Greedy} \\
\hline
Seq2Seq & N/A &  0.760 \\
\hline
RNN-DOM-Seq2Seq & 67.8\% & 0.797\\
\hline
En-DOM-Seq2Seq & {\bf 77.57\%} & {\bf 0.801} \\
\end{tabular}
\caption{Results for evaluation of Seq2seq, RNN based DOM-Seq2Seq (RNN-DOM-Seq2Seq), and Ensemble based DOM-Seq2Seq model (En-DOM-Seq2Seq)}
\label{tab:eval}
\end{table} 
\section{Conclusion}
We proposed an original technique for building a domain-aware chat system that can generate more relevant responses. We put forth two models to achieve this: an ensemble that considers a fixed number of previous domains and the current utterance to generate an appropriate response, and an RNN based classifier that models the switching of domains during a conversation. The methods described here can be extended to model a user's conversation pattern (domain shifts) and personalize the responses of the chat system.

\Urlmuskip=0mu plus 1mu
\bibliography{dom_seq2seq}

\begin{thebibliography}{}
\expandafter\ifx\csname natexlab\endcsname\relax\def\natexlab#1{#1}\fi

\bibitem[{Bahdanau et~al.(2014)Bahdanau, Cho, and
  Bengio}]{DBLP:journals/corr/BahdanauCB14}
Dzmitry Bahdanau, Kyunghyun Cho, and Yoshua Bengio. 2014.
\newblock \href{http://arxiv.org/abs/1409.0473}{Neural machine translation by
  jointly learning to align and translate}.
\newblock {\em CoRR\/} abs/1409.0473.
\newblock
  \href{http://arxiv.org/abs/1409.0473}{http://arxiv.org/abs/1409.0473}.

\bibitem[{Blei et~al.(2003)Blei, Ng, and Jordan}]{Blei:2003:LDA:944919.944937}
David~M. Blei, Andrew~Y. Ng, and Michael~I. Jordan. 2003.
\newblock \href{http://dl.acm.org/citation.cfm?id=944919.944937}{Latent
  dirichlet allocation}.
\newblock {\em J. Mach. Learn. Res.\/} 3:993--1022.
\newblock
  \href{http://dl.acm.org/citation.cfm?id=944919.944937}{http://dl.acm.org/citation.cfm?id=944919.944937}.

\bibitem[{Liu et~al.(2016)Liu, Lowe, Serban, Noseworthy, Charlin, and
  Pineau}]{DBLP:conf/emnlp/LiuLSNCP16}
Chia{-}Wei Liu, Ryan Lowe, Iulian Serban, Michael Noseworthy, Laurent Charlin,
  and Joelle Pineau. 2016.
\newblock \href{http://aclweb.org/anthology/D/D16/D16-1230.pdf}{How {NOT} to
  evaluate your dialogue system: An empirical study of unsupervised evaluation
  metrics for dialogue response generation}.
\newblock In Jian Su, Xavier Carreras, and Kevin Duh, editors, {\em Proceedings
  of the 2016 Conference on Empirical Methods in Natural Language Processing,
  {EMNLP} 2016, Austin, Texas, USA, November 1-4, 2016\/}. The Association for
  Computational Linguistics, pages 2122--2132.
\newblock
  \href{http://aclweb.org/anthology/D/D16/D16-1230.pdf}{http://aclweb.org/anthology/D/D16/D16-1230.pdf}.

\bibitem[{Pennington et~al.(2014)Pennington, Socher, and
  Manning}]{pennington2014glove}
Jeffrey Pennington, Richard Socher, and Christopher~D. Manning. 2014.
\newblock \href{http://www.aclweb.org/anthology/D14-1162}{Glove: Global vectors
  for word representation}.
\newblock In {\em Empirical Methods in Natural Language Processing (EMNLP)\/}.
  pages 1532--1543.
\newblock
  \href{http://www.aclweb.org/anthology/D14-1162}{http://www.aclweb.org/anthology/D14-1162}.

\bibitem[{Serban et~al.(2016)Serban, Sordoni, Bengio, Courville, and
  Pineau}]{DBLP:conf/aaai/SerbanSBCP16}
Iulian~Vlad Serban, Alessandro Sordoni, Yoshua Bengio, Aaron~C. Courville, and
  Joelle Pineau. 2016.
\newblock \href{http://www.aaai.org/ocs/index.php/AAAI/AAAI16/
  paper/view/11957}{Building end-to-end dialogue systems using generative
  hierarchical neural network models}.
\newblock In Dale Schuurmans and Michael~P. Wellman, editors, {\em Proceedings
  of the Thirtieth {AAAI} Conference on Artificial Intelligence, February
  12-17, 2016, Phoenix, Arizona, {USA.}\/}. {AAAI} Press, pages 3776--3784.
\newblock \href{http://www.aaai.org/ocs/index.php/AAAI/AAAI16/
  paper/view/11957}{http://www.aaai.org/ocs/index.php/AAAI/AAAI16/
  paper/view/11957}.

\bibitem[{Serban et~al.(2017)Serban, Sordoni, Lowe, Charlin, Pineau, Courville,
  and Bengio}]{DBLP:conf/aaai/SerbanSLCPCB17}
Iulian~Vlad Serban, Alessandro Sordoni, Ryan Lowe, Laurent Charlin, Joelle
  Pineau, Aaron~C. Courville, and Yoshua Bengio. 2017.
\newblock \href{http://aaai.org/ocs/index.php/AAAI/AAAI17/paper /view/14567}{A
  hierarchical latent variable encoder-decoder model for generating dialogues}.
\newblock In  \cite{DBLP:conf/aaai/2017}, pages 3295--3301.
\newblock \href{http://aaai.org/ocs/index.php/AAAI/AAAI17/paper
  /view/14567}{http://aaai.org/ocs/index.php/AAAI/AAAI17/paper /view/14567}.

\bibitem[{Singh and Markovitch(2017)}]{DBLP:conf/aaai/2017}
Satinder~P. Singh and Shaul Markovitch, editors. 2017.
\newblock {\em Proceedings of the Thirty-First {AAAI} Conference on Artificial
  Intelligence, February 4-9, 2017, San Francisco, California, {USA}\/}. {AAAI}
  Press.
\newblock \href{http://www.aaai.org/Library/AAAI/
  aaai17contents.php}{http://www.aaai.org/Library/AAAI/ aaai17contents.php}.

\bibitem[{Sutskever et~al.(2014)Sutskever, Vinyals, and
  Le}]{DBLP:journals/corr/SutskeverVL14}
Ilya Sutskever, Oriol Vinyals, and Quoc~V Le. 2014.
\newblock
  \href{http://papers.nips.cc/paper/5346-sequence-to-sequence-learning-with-neural-networks.pdf}{Sequence
  to sequence learning with neural networks}.
\newblock In Z.~Ghahramani, M.~Welling, C.~Cortes, N.~D. Lawrence, and K.~Q.
  Weinberger, editors, {\em Advances in Neural Information Processing Systems
  27\/}. Curran Associates, Inc., pages 3104--3112.
\newblock
  \href{http://papers.nips.cc/paper/5346-sequence-to-sequence-learning-with-neural-networks.pdf}{http://papers.nips.cc/paper/5346-sequence-to-sequence-learning-with-neural-networks.pdf}.

\bibitem[{Vinyals and Le(2015)}]{DBLP:journals/corr/VinyalsL15}
Oriol Vinyals and Quoc~V. Le. 2015.
\newblock \href{http://arxiv.org/abs/1506.05869}{A neural conversational
  model}.
\newblock {\em CoRR\/} abs/1506.05869.
\newblock
  \href{http://arxiv.org/abs/1506.05869}{http://arxiv.org/abs/1506.05869}.

\bibitem[{Xing et~al.(2017)Xing, Wu, Wu, Liu, Huang, Zhou, and
  Ma}]{DBLP:conf/aaai/XingWWLHZM17}
Chen Xing, Wei Wu, Yu~Wu, Jie Liu, Yalou Huang, Ming Zhou, and Wei{-}Ying Ma.
  2017.
\newblock \href{http://aaai.org/ocs/index.php/AAAI/AAAI17/paper/
  view/14563}{Topic aware neural response generation}.
\newblock In  \cite{DBLP:conf/aaai/2017}, pages 3351--3357.
\newblock \href{http://aaai.org/ocs/index.php/AAAI/AAAI17/paper/
  view/14563}{http://aaai.org/ocs/index.php/AAAI/AAAI17/paper/ view/14563}.

\end{thebibliography}
\bibliographystyle{acl_natbib}

\end{document}